\documentclass[11pt,a4paper]{article}
\usepackage[hyperref]{emnlp-ijcnlp-2019}
\usepackage{times}
\usepackage{latexsym}
\usepackage{paralist}
\usepackage{booktabs}
\usepackage{enumitem}
\usepackage{adjustbox}
\usepackage{subcaption}
\setitemize{noitemsep,topsep=0pt,parsep=0pt,partopsep=0pt}
\setenumerate{noitemsep,topsep=0pt,parsep=0pt,partopsep=0pt}

\usepackage{multirow}

\usepackage{graphicx}

\usepackage{url}

\aclfinalcopy 

\title{Revealing the Dark Secrets of BERT}

\author{
    Olga Kovaleva, Alexey Romanov, Anna Rogers, Anna Rumshisky  \\
    Department of Computer Science \\
    University of Massachusetts Lowell  \\
    Lowell, MA 01854\\ 
    {\tt \{okovalev,arum,aromanov\}@cs.uml.edu}
}

\date{}

\begin{document}
\maketitle
\begin{abstract}
  BERT-based architectures currently give state-of-the-art performance on many NLP tasks, but little is known about the exact mechanisms that contribute to its success.
  In the current work, we focus on the interpretation of self-attention, which is one of the fundamental underlying components of BERT. Using a subset of GLUE tasks and a set of handcrafted features-of-interest, we propose the methodology and carry out a qualitative and quantitative analysis of the information encoded by the individual BERT's heads.
  Our findings suggest that there is a limited set of attention patterns that are repeated across different heads, indicating the overall model overparametrization.  
  While different heads consistently use the same attention patterns, they have varying impact on performance across different tasks.  We show that manually disabling attention in certain heads leads to a performance improvement over the regular fine-tuned BERT models.

\end{abstract}

\section{Introduction}

Over the past year, models based on the Transformer architecture \cite{vaswani2017attention} have become the de-facto standard for state-of-the-art performance on many natural language processing (NLP) tasks \cite{radford2018improving, devlin2018bert}.
%
%
Their key feature is the self-attention mechanism that provides an alternative to conventionally used recurrent neural networks (RNN). 
%
%

One of the most popular Transformer-based models is BERT, which learns text representations using a bi-directional Transformer encoder pre-trained on the language modeling task \cite{devlin2018bert}.
BERT-based architectures have produced new state-of-the-art performance on a range of NLP tasks of different nature, domain, and complexity, including question answering, sequence tagging, sentiment analysis, and inference.
State-of-the-art performance is usually obtained by fine-tuning the pre-trained model on the specific task.
%
%
In particular, BERT-based models are currently dominating the leaderboards for SQuAD\footnote{\url{https://rajpurkar.github.io/SQuAD-explorer/}} \cite{rajpurkar2016squad} and GLUE benchmarks\footnote{\url{https://gluebenchmark.com/leaderboard}} \cite{wang2018glue}. 

However, the exact mechanisms that contribute to the BERT's outstanding performance still remain unclear. We address this problem through selecting a set of linguistic features of interest and conducting a series of experiments that aim to provide insights about how well these features are captured by BERT. This paper makes the following contributions:
\begin{itemize}
    \item We propose a methodology and offer the first detailed analysis of BERT's capacity to capture different kinds of linguistic information by encoding it in its self-attention weights.
    \item We present evidence of BERT's overparametrization and suggest a counter-intuitive yet frustratingly simple way of improving its performance, showing absolute gains of up to 3.2\%.
\end{itemize}

\section{Related work}

There have been several recent attempts to assess BERT's ability to capture structural properties of language.
\citet{goldberg2019assessing} demonstrated that BERT consistently assigns higher scores to the correct verb forms as opposed to the incorrect one in a masked language modeling task, suggesting some ability to model subject-verb agreement. \citet{jawahar:hal-02131630} extended this work to using multiple layers and tasks, supporting the claim that BERT's intermediate layers capture rich linguistic information.
On the other hand, \citet{tran2018importance} 
concluded that LSTMs generalize to longer sequences better, and are more robust with respect to agreement distractors, compared to Transformers. 

\citet{liu2019linguistic} investigated the transferability of contextualized word representations to a number of probing tasks requiring linguistic knowledge. Their findings suggest that (a) the middle layers of Transformer-based architectures are the most transferable to other tasks, and (b) higher layers of Transformers are not as task specific as the ones of RNNs.
\citet{tang2018self} argued that models using self-attention outperform CNN-  and RNN-based models on a word sense disambiguation task due to their ability to extract semantic features from text. \citet{voita2019analyzing} analyzed the original Transformer model on a translation task and found out that only a small subset of heads is important for the given task, but these heads have interpretable linguistic functions.

Our work contributes to the above discussion, but rather than examining representations extracted from different layers, we focus on the understanding of the self-attention mechanism itself, since it is the key feature of Transformer-based models.

Another research direction that is relevant to our work is neural network pruning. \citet{frankle2018lottery} showed that widely used complex architectures suffer from overparameterization, and can be significantly reduced in size without a loss in performance. 
\citet{goldberg2019assessing} observed that the smaller version of BERT achieves better scores on a number of syntax-testing experiments than the larger one. \citet{adhikari2019rethinking} questioned the necessity of computation-heavy neural networks, proving that a simple yet carefully tuned BiLSTM without attention achieves the best or at least competitive results compared to more complex architectures on the document classification task. \citet{wu2019pay} presented more evidence of unnecessary complexity of the self-attention mechanism, and proposed a more lightweight and scalable dynamic convolution-based architecture that outperforms the self-attention baseline. \citet{michel2019sixteen} demonstrated that some layers in Transformer can be reduced down to a single head without significant degradation of model performance.
These studies suggest a potential direction for future research, and are in good accordance with our observations.


\section{Methodology}

We pose the following research questions: 
\begin{enumerate}
    \item What are the common attention patterns, how do they change during fine-tuning, and how does that impact the performance on a given task? (Sec. \ref{sec:patterns}, \ref{sec:fine-tuning})
    \item What linguistic knowledge is encoded in self-attention weights of the fine-tuned models and what portion of it comes from the pre-trained BERT? (Sec. \ref{sec:fn}, \ref{sec:vert_attention},  \ref{sec:cross_attention})
    \item How different are the self-attention patterns of different heads, and how important are they for a given task? (Sec. \ref{sec:disabling})
\end{enumerate}

The answers to these questions come from a series of experiments with the basic pre-trained or the fine-tuned BERT models, as will be discussed below.
All the experiments with the pre-trained BERT were conducted using the model provided with the PyTorch implementation of BERT (bert-base-uncased, 12-layer, 768-hidden, 12-heads, 110M parameters)\footnote{\url{https://github.com/huggingface/pytorch-pretrained-BERT}}. We chose this smaller version of BERT because it shows competitive, if not better, performance while having fewer layers and heads, which makes it more interpretable.

We use the following subset of GLUE tasks \cite{wang2018glue} for fine-tuning: 
\begin{itemize}
    \item \textit{MRPC}: the Microsoft Research Paraphrase Corpus~\cite{dolan2005automatically}
    \item \textit{STS-B}: the Semantic Textual Similarity Benchmark~\cite{cer2017semeval}
    \item \textit{SST-2}: the Stanford Sentiment Treebank, two-way classification~\cite{socher2013recursive}
    \item \textit{QQP}: the Quora Question Pairs dataset
    \item \textit{RTE}: the Recognizing Textual Entailment datasets
    \item \textit{QNLI}: Question-answering NLI based on the Stanford Question Answering Dataset~\cite{rajpurkar2016squad}
    \item \textit{MNLI}: the Multi-Genre Natural Language Inference Corpus, matched section~\cite{williams2018broad}
\end{itemize}

Please refer to the original GLUE paper for details on the QQP and RTE datasets~\cite{wang2018glue}. We excluded two tasks: CoLa and the Winograd Schema Challenge. The latter is excluded due to the small size of the dataset. As for CoLa (the task of predicting linguistic acceptability judgments), GLUE authors report that the human performance is only 66.4, which is explained by the problems with the underlying methodology \cite{Schutze_1996_Empirical_Base_of_Linguistics_Grammaticality_Judgments_and_Linguistic_Methodology}. Note also that CoLa is not included in the upcoming version of GLUE \cite{WangPruksachatkunEtAl_2019_SuperGLUE_Stickier_Benchmark_for_General-Purpose_Language_Understanding_Systems}.
All fine-tuning experiments follow the parameters reported in the original study (a batch size of 32 and 3 epochs, see~\newcite{devlin2018bert}).

In all these experiments, for a given input, we extract self-attention weights for each head in every layer. This results in a 2D float array of shape $L\times L$, where $L$ is the length of an input sequence. We will refer to such arrays as \textit{self-attention maps}. Analysis of individual self-attention maps allows us to determine which target tokens are attended to the most as the input is processed token by token. 
We use these experiments to analyze how BERT processes different kinds of linguistic information, including the processing of different parts of speech (nouns, pronouns, and verbs), syntactic roles (objects, subjects), semantic relations, and negation tokens.


\begin{figure*}[ht]
  \centering
  \includegraphics[width=\linewidth]{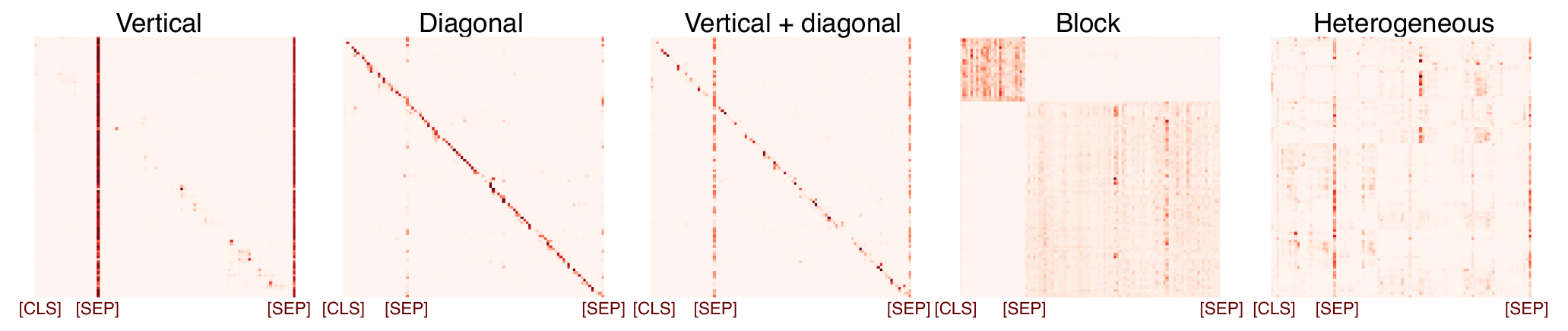}
  \caption{Typical self-attention classes used for training a neural network. Both axes on every image represent BERT tokens of an input example, and colors denote absolute attention weights (darker colors stand for greater weights). The first three types are most likely associated with language model pre-training, while the last two potentially encode semantic and syntactic information.}
  \label{fig:att_types}
\end{figure*}

\section{Experiments}
In this section, we present the experiments conducted to address the above research questions.  

\subsection{BERT's self-attention patterns} \label{sec:patterns}


Manual inspection of self-attention maps for both basic pre-trained and fine-tuned BERT models suggested that there is a limited set of self-attention map types that are repeatedly encoded across different heads. Consistently with previous observations\footnote{\url{https://towardsdatascience.com/deconstructing-bert-distilling-6-patterns-\\from-100-million-parameters-b49113672f77}}, we identified five frequently occurring patterns, examples of which are shown in \autoref{fig:att_types}:
\begin{itemize}
    \item \textit{Vertical}: mainly corresponds to  attention to special BERT tokens \textit{[CLS]} and \textit{[SEP]} which serve as delimiters between individual chunks of BERT's inputs;
    \item \textit{Diagonal}: formed by the attention to the previous/following tokens;
    \item \textit{Vertical+Diagonal}: a mix of the previous two types,
    \item \textit{Block}: intra-sentence attention for the tasks with two distinct sentences (such as, for example, RTE or MRPC),
    \item \textit{Heterogeneous}: highly variable depending on the specific input and cannot be characterized by a distinct structure.
\end{itemize}
Note that, because the \textit{Heterogeneous} category contains patterns not included in the other four categories, our constructed list of classes is exhaustive.

Whereas the attention to the special tokens is important for cross-sentence reasoning, and the attention to the previous/following token comes from language model pre-training, we hypothesize that the last of the listed types is more likely to capture interpretable linguistic features, necessary for language understanding.

To get a rough estimate of the percentage of attention heads that may capture linguistically interpretable information, we manually annotated around 400 sample self-attention maps as belonging to one of the five classes. The self-attention maps were obtained by feeding random input examples from selected tasks into the corresponding fine-tuned BERT model.  This produced a somewhat unbalanced dataset, in which the ``Vertical'' class accounted for 30\% of all samples.  We then trained a convolutional neural network with 8 convolutional layers and ReLU activation functions to classify input maps into one of these classes.  This model achieved the F1 score of 0.86 on the annotated dataset.  We used this classifier to estimate the proportion of different self-attention patterns for the target GLUE tasks using up to 1000 examples (where available) from each validation set. 



\begin{figure}[ht]
  \centering
  \includegraphics[width=\linewidth]{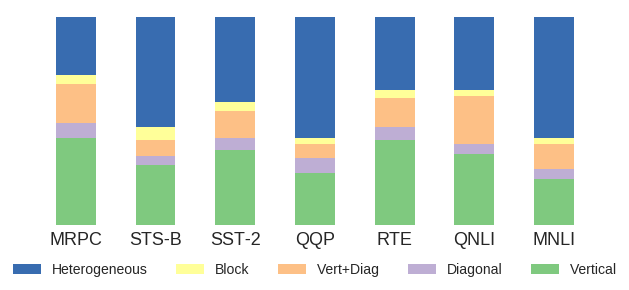}
  \caption{Estimated percentages of the identified self-attention classes for each of the selected GLUE tasks. }
  \label{fig:attention_by_dataset}
\end{figure}

\paragraph*{Results}
\autoref{fig:attention_by_dataset} shows that the self-attention map types described above are consistently repeated across different heads and tasks. While a large portion of encoded information corresponds to attention to the previous/following token, to the special tokens, or a mixture of the two (the first three classes), the estimated upper bound on all heads in the ``Heterogeneous'' category (i.e. the ones that \textit{could} be informative) varies from 32\% (MRPC) to 61\% (QQP) depending on the task. 

We would like to emphasize that this only gives the upper bound on the percentage of attention heads that could potentially capture meaningful structural information beyond adjacency and separator tokens.



\subsection{Relation-specific heads in BERT}
\label{sec:fn}

In this experiment, our goal was to understand whether different syntactic and semantic relations are captured by self-attention patterns. While a large number of such relations could be investigated, we chose to examine semantic role relations defined in frame semantics, since they can be viewed as being at the intersection of syntax and semantics. Specifically, we focused on whether BERT captures FrameNet's relations between frame-evoking lexical units (predicates) and core frame elements \cite{baker1998berkeley}, and whether the links between them produce higher attention weights in certain specific heads.  We used pre-trained BERT in these experiments.



\begin{figure*}
  \centering
  \includegraphics[width=0.9\linewidth]{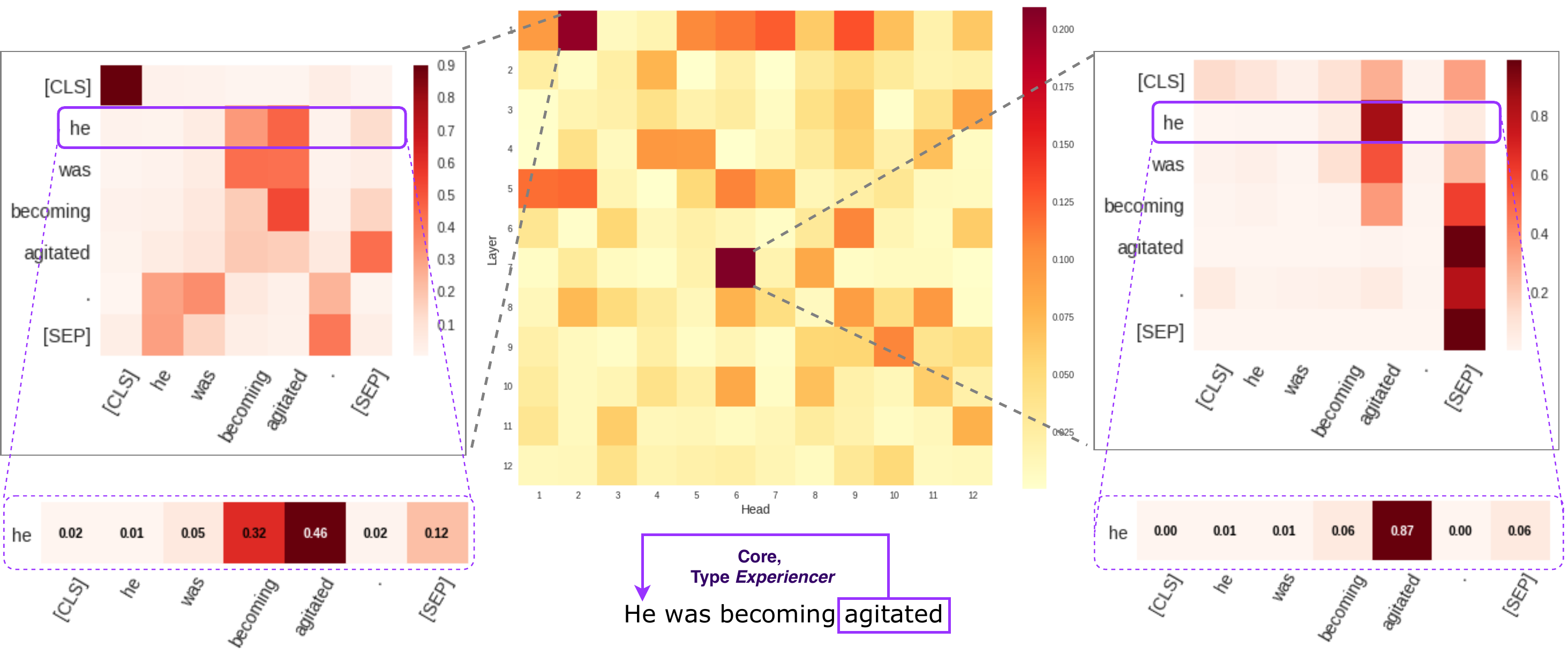}
  \caption{Detection of pre-trained BERT's heads that encode information correlated to semantic links in the input text. Two heads (middle) demonstrate their ability to capture semantic relations. Note that the heatmap in the middle is obtained through averaging of all the individual input example maps. For one random annotated FrameNet example (bottom) full attention maps with a zoom in the target token attention distribution are shown (leftmost and rightmost).}
  \label{fig:framenet_results}
\end{figure*}

\begin{figure}[!ht]
  \centering
  \includegraphics[width=0.8\linewidth]{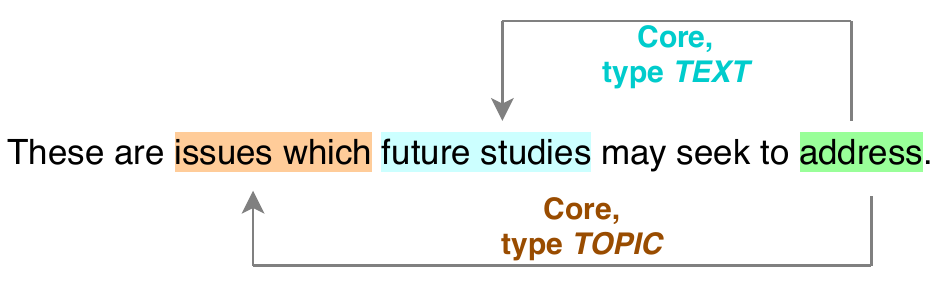}
  \caption{FrameNet annotation example for the ``address'' lexical unit with two core frame elements of different types annotated.}
  \label{fig:framenet}
\end{figure}

The data for this experiment comes from FrameNet \cite{baker1998berkeley}, a database that contains frame annotations for example sentences for different lexical units.
Frame elements correspond to semantic roles for a given frame, for example, ``buyer", ``seller", and ``goods'' for the ``Commercial\_transaction" frame evoked by the words ``sell'' and ``spend'' or ``topic'' and ``text'' for the ``Scrutiny'' semantic frame evoked by the verb ``address''. \autoref{fig:framenet} shows an example of such annotation.


We extracted sample sentences for every lexical unit in the database and identified the corresponding core frame elements. Annotated elements in FrameNet may be rather long, 
so we considered only the sentences with frame elements of 3 tokens or less. Since each sentence is annotated only for one frame, semantic links from other frames can exist between unmarked elements. We therefore filter out all the sentences longer than 12 tokens, since shorter sentences are less likely to evoke multiple frames.

To establish whether BERT attention captures semantic relations that \textit{do not} simply correspond to the previous/following token, we exclude sentences where the linked objects are less than two tokens apart. This leaves us with 473 annotated sentences.

For each of these sentences, we obtain pre-trained BERT's attention weights for each of the 144 heads. For every head, we return the maximum absolute attention weight among those token pairs that correspond to the annotated semantic link contained within a given sentence. We then average the derived scores over all the collected examples. This strategy allows us to identify the heads that prioritize the features correlated with frame-semantic relations within a sentence.

\paragraph*{Results}
The heatmap of averaged attention scores over all collected examples (\autoref{fig:framenet_results}) suggests that 2 out of 144 heads tend to attend to the parts of the sentence that FrameNet annotators identified as core elements of the same frame. The maximum attention weights averaged over all data examples for these identified heads account for $0.201$ and $0.209$, which are greater than a 99-th percentile of the distribution of values for all heads. \autoref{fig:framenet_results} shows an example of this attention pattern for these two heads. Both show high attention weight for ``he'' while processing ``agitated'' in the sentence ``He was becoming agitated" (the frame ``Emotion\_directed'').

We interpret these results as limited evidence that certain types of linguistic relations may be captured by self-attention patterns in specialized BERT heads. A wider range of relations remains to be investigated.



  
\begin{figure*}[ht]
  \centering
  \includegraphics[width=\linewidth]{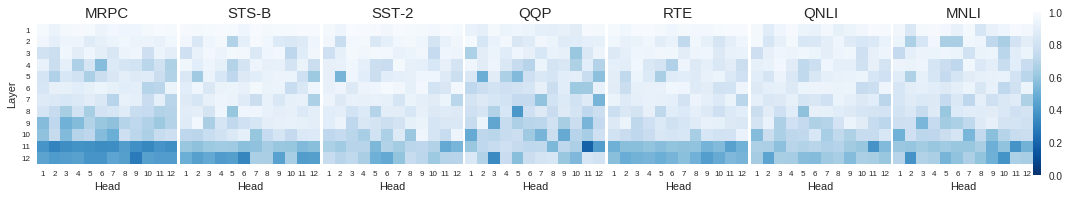}
  \caption{Per-head cosine similarity between pre-trained BERT's and fine-tuned BERT's self-attention maps for each of the selected GLUE tasks, averaged over validation dataset examples. Darker colors correspond to greater differences.}
  \label{fig:cosine}
\end{figure*}

\subsection{Change in self-attention patterns after fine-tuning} \label{sec:fine-tuning}


Fine-tuning has a huge effect on performance, and this section attempts to find out why. 
To study 
how attention per head changes on average for each of the target GLUE tasks, we calculate cosine similarity between pre-trained and fine-tuned BERT's flattened arrays of attention weights. We average the derived similarities over all the development set examples\footnote{If the number of development data examples for a given task exceeded 1000 (QQP, QNLI, MNLI, STS-B), we randomly sampled 1000 examples.}. To evaluate contribution of pre-trained BERT to overall performance on the tasks, we consider two configurations of weights initialization, namely, pre-trained BERT weights and weights randomly sampled from normal distribution. 

\begin{figure*}
  \includegraphics[width=\linewidth]{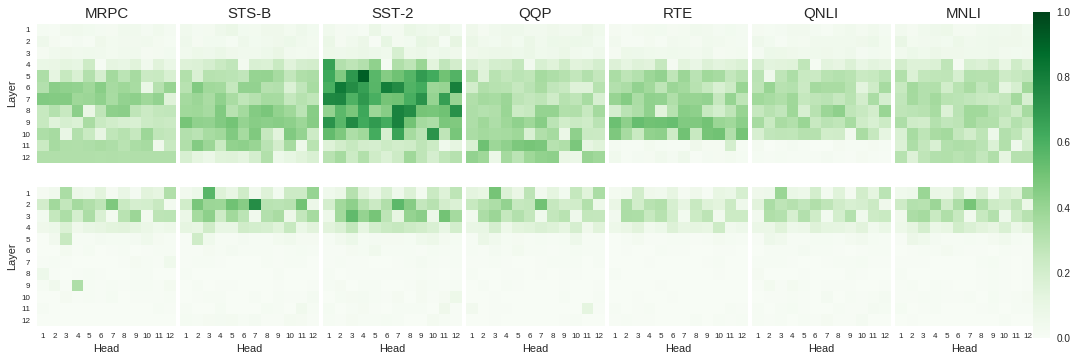}
  \caption{Per-task attention weights to the \textit{[SEP]} (top row) and the \textit{[CLS]} (bottom row) tokens averaged over input sequences' lengths and over dataset examples. Darker colors correspond to greater absolute weights.}
  \label{fig:special_tokens}
\end{figure*}

\begin{figure*}
  \includegraphics[width=\linewidth]{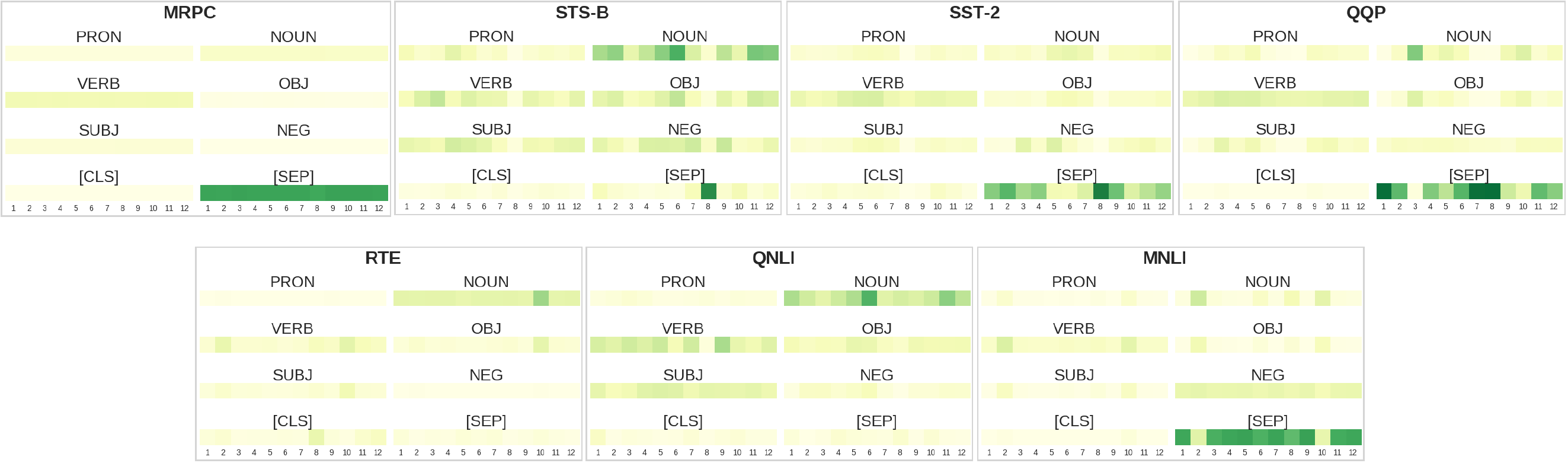}
  \caption{Per-task attention weights corresponding to the \textit{[CLS]} token averaged over input sequences' lengths and over dataset examples, and extracted from the final layer. Darker colors correspond to greater absolute weights.}
  \label{fig:cls}
\end{figure*}

\begin{table}[!hb]
\centering
\footnotesize
\begin{adjustbox}{max width=1\linewidth}
\begin{tabular}{@{}lrrrrr@{}}
\toprule
\multirow{2}{*}{\textbf{Dataset}} & \multirow{2}{*}{\textbf{Pre-trained}} & \multicolumn{2}{c}{\textbf{Fine-tuned, initialized with}} &  \multirow{2}{*}{\textbf{Metric}} &
\multirow{2}{*}{\textbf{Size}}\\
 & & \textbf{normal distr.} & \textbf{pre-trained} &
 \\
\midrule
MRPC    & 0/31.6    & 81.2/68.3 & 87.9/82.3 & F1/Acc & 5.8K\\
STS-B   & 33.1      & 2.9       & 82.7      & Acc    & 8.6K\\
SST-2   & 49.1      & 80.5      & 92        & Acc    & 70K\\
QQP     & 0/60.9    & 0/63.2    & 65.2/78.6 & F1/Acc & 400K\\
RTE     & 52.7      & 52.7      & 64.6      & Acc    & 2.7K\\
QNLI    & 52.8      & 49.5      & 84.4      & Acc    & 130K\\
MNLI-m  & 31.7      &  61.0     & 78.6      & Acc    & 440K\\ \bottomrule
\end{tabular}
\end{adjustbox}
\caption{GLUE task performance of BERT models with different initialization. 
We report the scores on the validation, rather than test data, so these results differ from the original BERT paper.}
\label{tab:glue-results}
\end{table}

\paragraph*{Results}
\autoref{fig:cosine} shows that for all the tasks except QQP, it is the last two layers that undergo the largest changes compared to the pre-trained BERT model. At the same time, \autoref{tab:glue-results} shows that fine-tuned BERT outperforms pre-trained BERT by a significant margin on all the tasks (with an average of 35.9 points of absolute difference). This leads us to conclude that the last two layers encode task-specific features that are attributed to the gain of scores, while earlier layers capture more fundamental and low-level information used in fine-tuned models. 
BERT with weights initialized from normal distribution and further fine-tuned for a given task consistently produces lower scores than the ones achieved with pre-trained BERT. In fact, for some tasks (STS-B and QNLI), initialization with random weights yields worse performance than pre-trained BERT without fine-tuning. 

This suggests that pre-trained BERT does indeed contain linguistic knowledge that is helpful for solving these GLUE tasks. These results are consistent with similar studies, e.g., \citet{yosinski2014transferable}'s results on fine-tuning a convolutional neural network pre-trained on ImageNet or \citet{romanov2018lessons}'s results on transfer learning for medical natural language inference.

\subsection{Attention to linguistic features} \label{sec:vert_attention}

In this experiment, we investigate whether fine-tuning BERT for a given task creates self-attention patterns which emphasize specific linguistic features.  In this case, certain kinds of tokens may get high attention weights from all the other tokens in the sentence, 
producing vertical stripes on the corresponding attention maps~(\autoref{fig:att_types}).

To test this hypothesis we checked whether there are vertical stripe patterns corresponding to certain linguistically interpretable features, and to what extent such features are relevant for solving a given task. In particular, we investigated attention to nouns, verbs, pronouns, subjects, objects, and negation words\footnote{Our manually constructed list of negation words consisted of the following words \textit{neither, nor, not, never, none, don't, won't, didn't, hadn't, haven't, can't, isn't, wasn't, shouldn't, couldn't, nothing, nowhere.}}, and special BERT tokens across the tasks.


For every head, we compute the sum of self-attention weights assigned to the token of interest from each input token. Since the weights depend on the number of tokens in the input sequence, this sum is normalized by sequence length. This allows us to aggregate the weights for this feature across different examples. 
If there are multiple tokens of the same type (e.g. several nouns or negations), we take the maximum value.  We disregard input sentences that do not contain a given feature. 

For each investigated feature, we calculate this aggregated attention score for each head in every layer and build a map in order to detect the heads potentially responsible for this feature.  We then compare the obtained maps to the ones derived using the pre-trained BERT model. This comparison enables us to determine if a particular feature is important for a specific task and whether it contributes to some tasks more than to others.

\paragraph*{Results}
Contrary to our initial hypothesis that the vertical attention pattern may be motivated by linguistically meaningful features, we found that it is associated predominantly, if not exclusively, with attention to \textit{[CLS]} and \textit{[SEP]} tokens (see Figure \ref{fig:special_tokens}. Note that the absolute \textit{[SEP]} weights for the SST-2 sentiment analysis task are greater than for other tasks, which is explained by the fact that there is only one sentence in the model inputs, i.e. only one \textit{[SEP]} token instead of two. There is also a clear tendency for earlier layers to pay attention to \textit{[CLS]} and for later layers to \textit{[SEP]}, and this trend is consistent across all the tasks. 
We did detect heads that paid increased (compared to the pre-trained BERT) attention to nouns and direct objects of the main predicates (on the MRPC, RTE and QQP tasks), and negation tokens (on the QNLI task), but the attention weights of such tokens were negligible compared to \textit{[CLS]} and \textit{[SEP]}. Therefore, we believe that the striped attention maps generally come from BERT pre-training tasks rather than from task-specific linguistic reasoning.

\subsection{Token-to-token attention} \label{sec:cross_attention}
To complement the experiments in Sec.~\ref{sec:vert_attention}
and~\ref{sec:fn}, in this section, we investigate the attention patterns between tokens in the same sentence, 
i.e. whether any of the tokens are particularly important while a \textit{given} token is being processed. We were interested specifically in the verb-subject relation and the noun-pronoun relation. Also, since BERT uses the representation of the \textit{[CLS]} token in the last layer to make the prediction, we used the features from the experiment in Sec.~\ref{sec:vert_attention} in order to check if they get higher attention weights while the model is processing the \textit{[CLS]} token.


\paragraph*{Results}
Our token-to-token attention experiments for detecting heads that prioritize noun-pronoun and verb-subject links resulted in a set of potential head candidates that coincided with diagonally structured attention maps. We believe that this happened due to the inherent property of English syntax where the dependent elements frequently appear close to each other, so it is difficult to distinguish such relations from the previous/following token attention coming from language model pre-training. 

Our investigation of attention distribution for the \textit{[CLS]} token in the output layer suggests that for most tasks, with the exception of STS-B, RTE and QNLI, the \textit{[SEP]} gets attended the most, as shown in~\autoref{fig:cls}.  Based on manual inspection, for the mentioned remaining tasks, the greatest attention weights correspond to the punctuation tokens, which are in a sense similar to \textit{[SEP]}.

\begin{figure*}[ht]
  \centering
  \includegraphics[width=\linewidth, trim={5cm 1cm 4.3cm 1cm},clip]{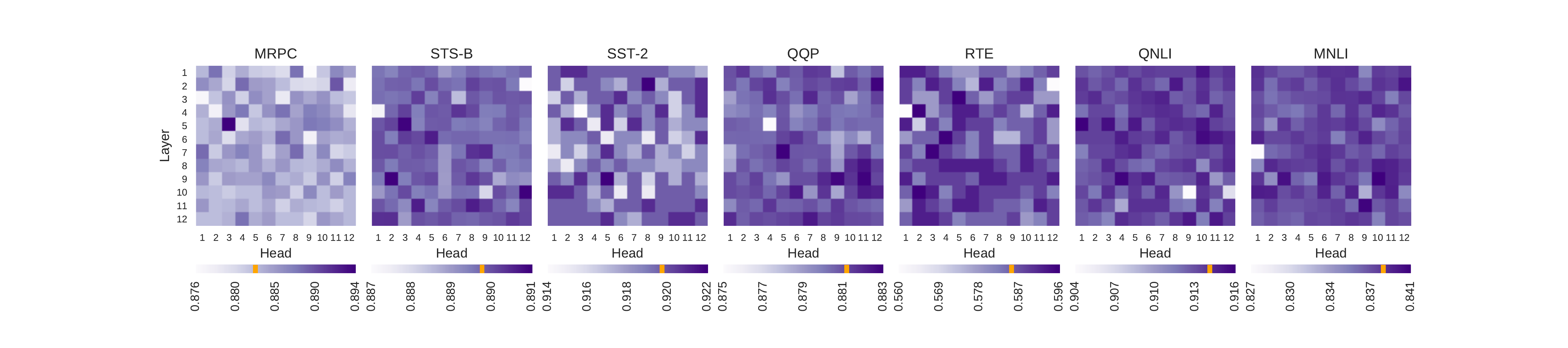}
  \caption{Performance of the model while disabling one head at a time. The orange line indicates the baseline performance with no disabled heads. Darker colors correspond to greater performance scores.}
  \label{fig:disable_heads_all}
\end{figure*}

\begin{figure*}[ht]
  \centering
  \includegraphics[width=0.7\linewidth, trim={5cm 1cm 2cm 1cm},clip]{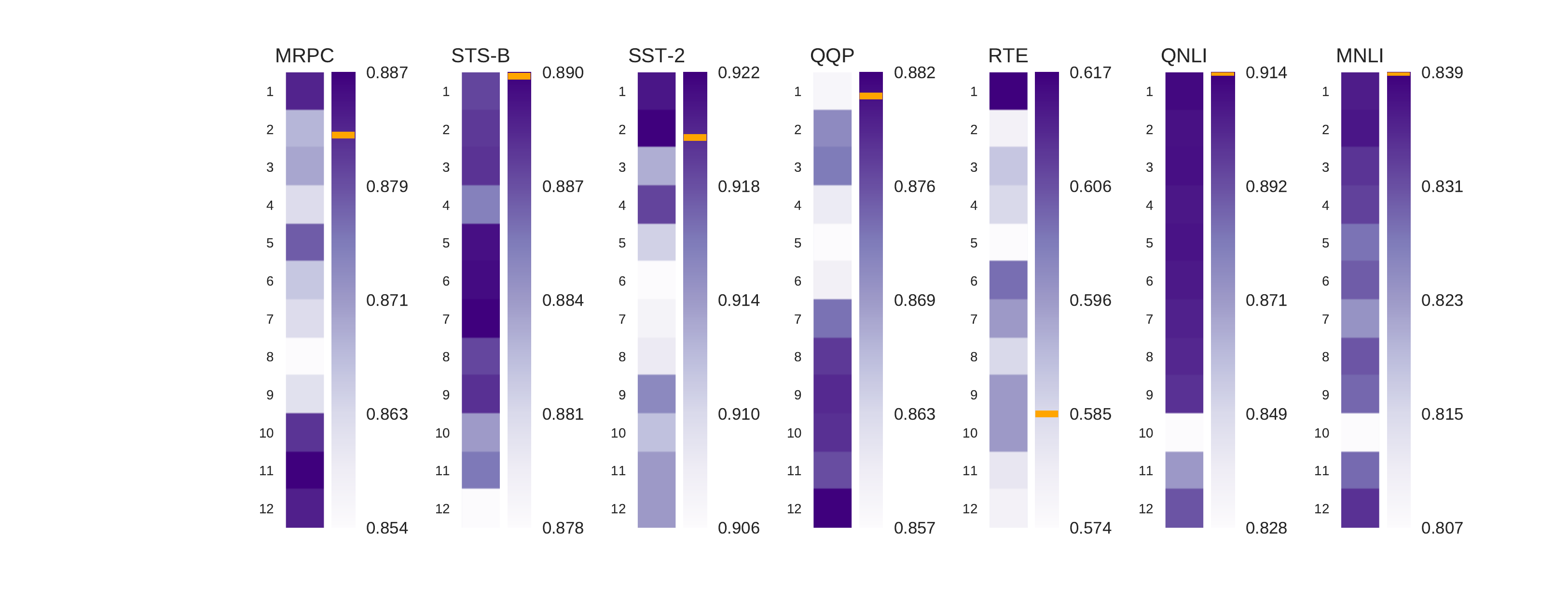}
  \caption{Performance of the model while disabling one layer (that is, all 12 heads in this layer) at a time. The orange line indicates the baseline performance with no disabled layers. Darker colors correspond to greater performance scores.}
  \label{fig:disable_layers}
\end{figure*}

\subsection{Disabling self-attention heads}
\label{sec:disabling}

Since there does seem to be a certain degree of specialization for different heads, we investigated the effects of disabling different heads in BERT and the resulting effects on  task performance. Since BERT relies heavily on the learned attention weights, we define disabling a head as modifying the attention values of a head to be constant $a = \frac{1}{L}$ for every token in the input sentence, where $L$ is the length of the sentence. Thus, every token receives the same attention, effectively disabling the learned attention patterns while maintaining the information flow of the original model. Note that by using this framework, we can disable an arbitrary number of heads, ranging from a single head per model to the whole layer or multiple layers.
\paragraph*{Results}


Our experiments suggest that certain heads have a detrimental effect on the overall performance of BERT, and this trend holds for all the chosen tasks. Unexpectedly, disabling some heads leads \textit{not} to a drop in accuracy, as one would expect, but to an increase in performance. This is effect is different across tasks and datasets. While disabling some heads improves the results, disabling the others hurts the results. However, it is important to note that across all tasks and datasets, disabling some heads leads to an increase in performance. The gain from disabling a single head is different for different tasks, ranging from the minimum absolute gain of 0.1\% for STS-B, to the maximum of 1.2\% for MRPC (see~\autoref{fig:disable_heads_all}). 
In fact, for some tasks, such as MRPC and RTE, disabling a \textit{random} head gives, on average, \textit{an increase} in performance. Furthermore, disabling a whole layer, that is, all 12 heads in a given layer, also improves the results. \autoref{fig:disable_layers} shows the resulting model performance on the target GLUE tasks when different layers are disabled. Notably, disabling the first layer in the RTE task gives a significant boost, resulting in an absolute performance gain of 3.2\%. However, effects of this operation vary across tasks, and for QNLI and MNLI, it produces a performance drop of up to -0.2\%.

\section{Discussion}
\label{sec:discussion}

In general, our results suggest that even the smaller base BERT model is significantly overparametrized. This is supported by the discovery of repeated self-attention patterns in different heads, as well as the fact that disabling both single and multiple heads is not detrimental to model performance and in some cases even improves it.


We found no evidence that attention patterns that are mappable onto core frame-semantic relations actually improve BERT's performance. 2 out of 144 heads that seem to be ``responsible" for these relations (see Section \ref{sec:fn}) do not appear to be important in any of the GLUE tasks: disabling of either one does not lead to a drop of accuracy. This implies that fine-tuned BERT does not rely on this piece of semantic information and prioritizes other features instead. 

For instance, we noticed that both STS-B and RTE fine-tuned models rely on attention in the same pair of heads (head 1 in the fourth layer, and head 12 in the second layer), as shown in Figure \ref{fig:disable_heads_all}. We manually checked the attention maps in those heads for a set of random inputs, and established that both of them have high weights for words that appear in both sentences of the input examples. This most likely means that word-by-word comparison of the two sentences provides a solid strategy of making a classification prediction for STS-B and RTE. We were not able to find a conceptually similar interpretation of heads important for other tasks.

\section{Conclusion}
In this work, we proposed a set of methods for analyzing self-attention mechanisms of BERT, 
comparing attention patterns for the pre-trained and fine-tuned versions of BERT. 


Our most surprising finding is that, although attention is the key BERT's underlying mechanism, the model can benefit from attention ``disabling". Moreover, we demonstrated that there is redundancy in the information encoded by different heads and the same patterns get consistently repeated regardless of the target task. We believe that these two findings together suggest a further direction for research on BERT interpretation, namely, model pruning and finding an optimal sub-architecture reducing data repetition.

One of the directions for future research would be to study self-attention patterns in different languages, especially verb-final langauges and those with free word order. It is possible that English has relatively lower variety of self-attention patterns, as the subject-predicate relation happens to coincide with the following-previous token pattern.

\section{Acknowledgments}
We thank David Donahue for the help with image annotation.
This project is funded in part by the NSF award number IIS-1844740.

\bibliography{emnlp-ijcnlp-2019}
\bibliographystyle{acl_natbib}

\clearpage
\appendix

\section{Examples of full self-attention maps}
In this section, we present examples of full self-attention maps of a set of fine-tuned models to provide a better illustration of different head patterns. All the maps are given for a randomly sampled example from a corresponding dataset.

\begin{figure}[ht]
\centering
\begin{minipage}{.5\textwidth}
  \centering
    \includegraphics[width=0.9\linewidth]{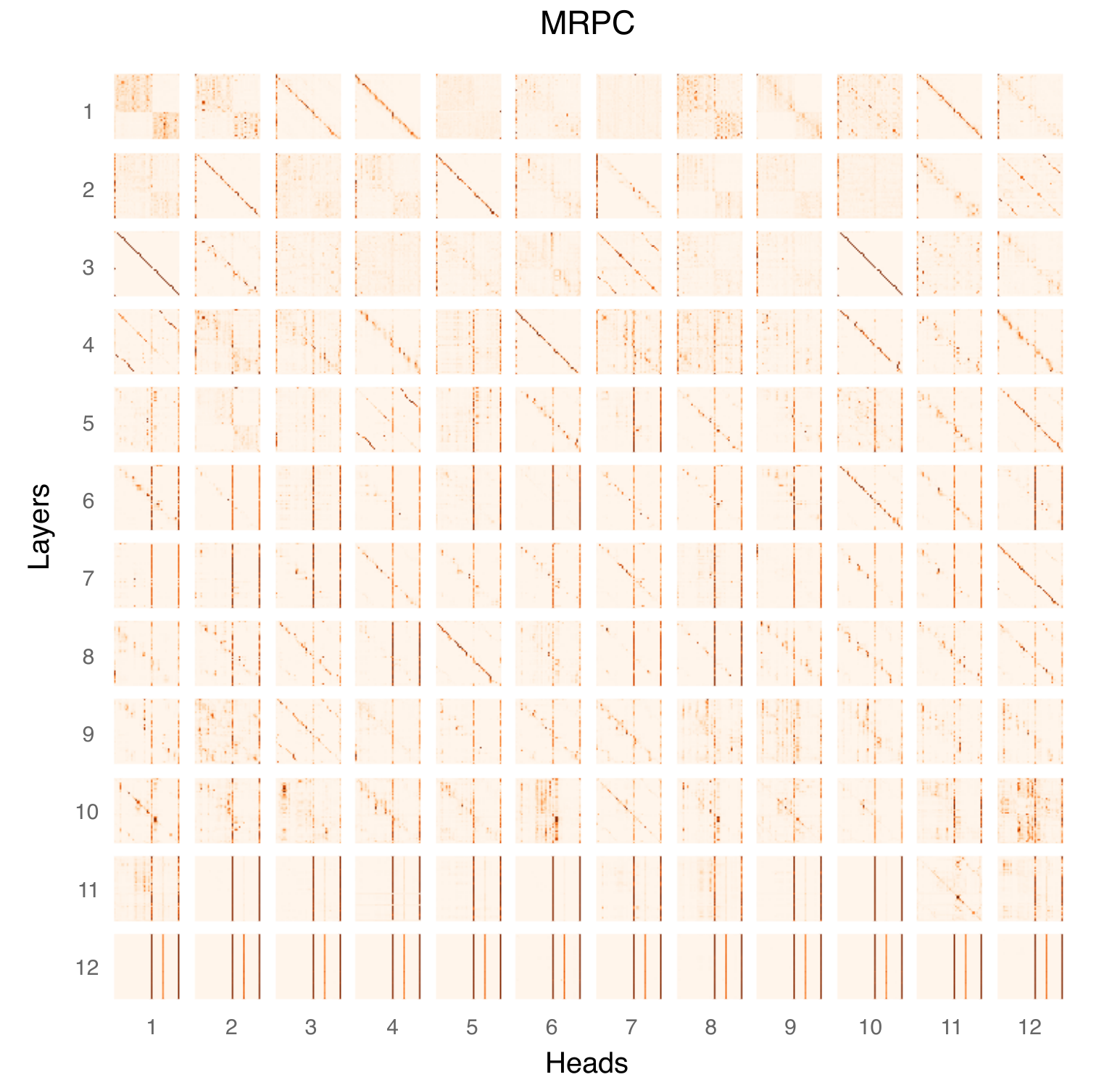}
\end{minipage}%
\begin{minipage}{.5\textwidth}
  \centering
    \includegraphics[width=0.9\linewidth]{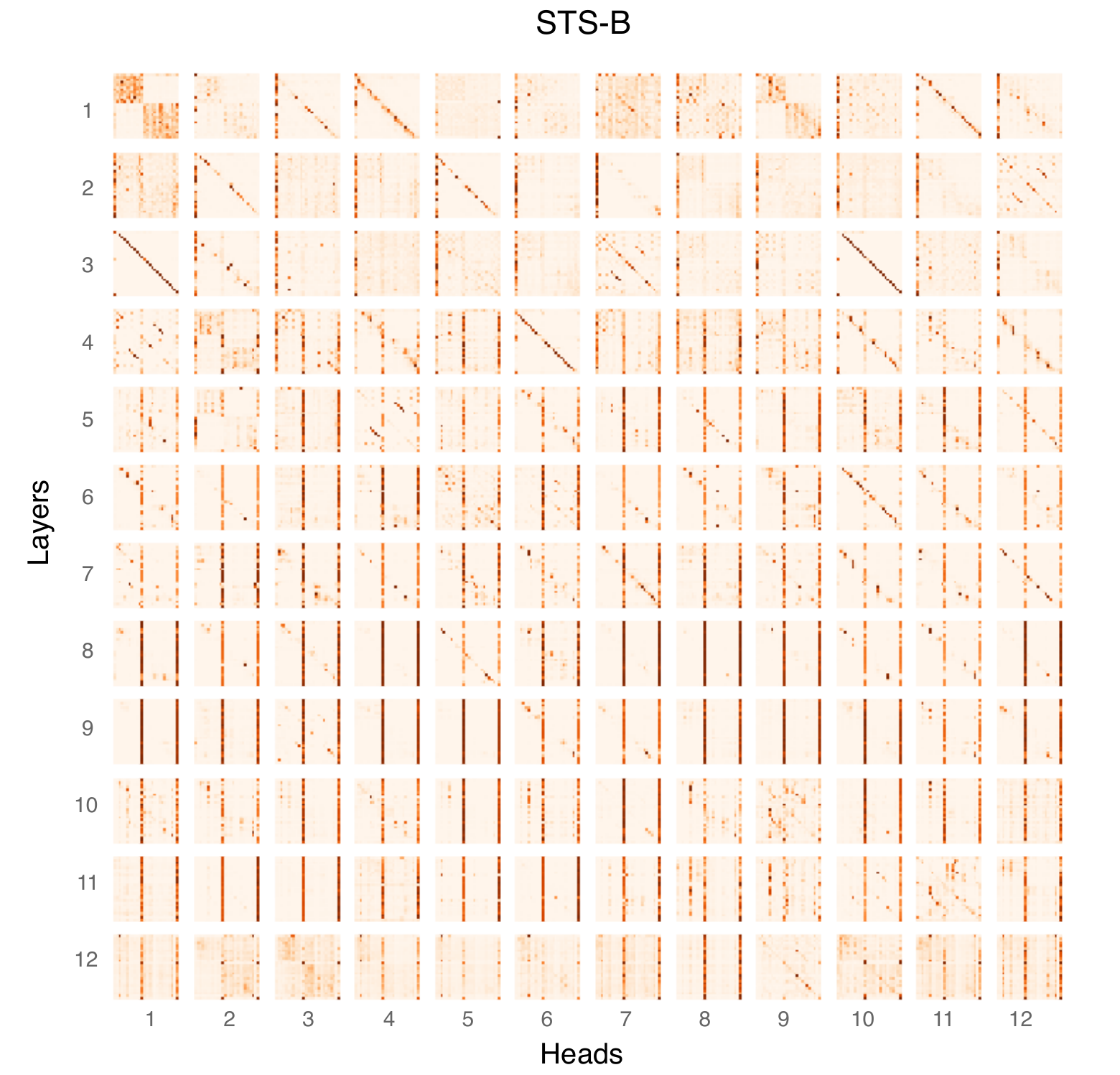}
\end{minipage}
\end{figure}

\begin{figure}[ht]
\centering
\begin{minipage}{.5\textwidth}
  \centering
    \includegraphics[width=0.9\linewidth]{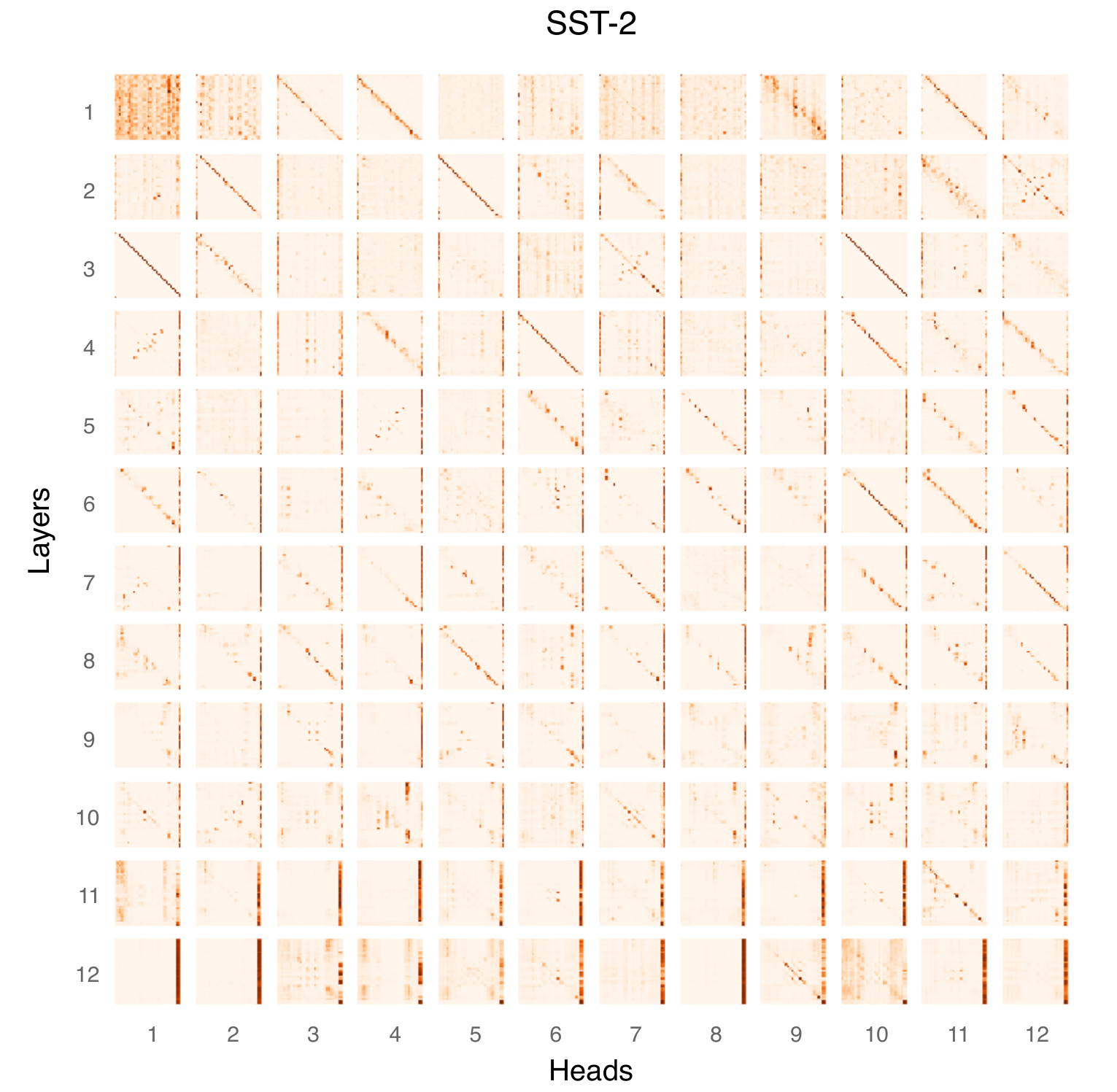}
\end{minipage}%
\begin{minipage}{.5\textwidth}
  \centering
    \includegraphics[width=0.9\linewidth]{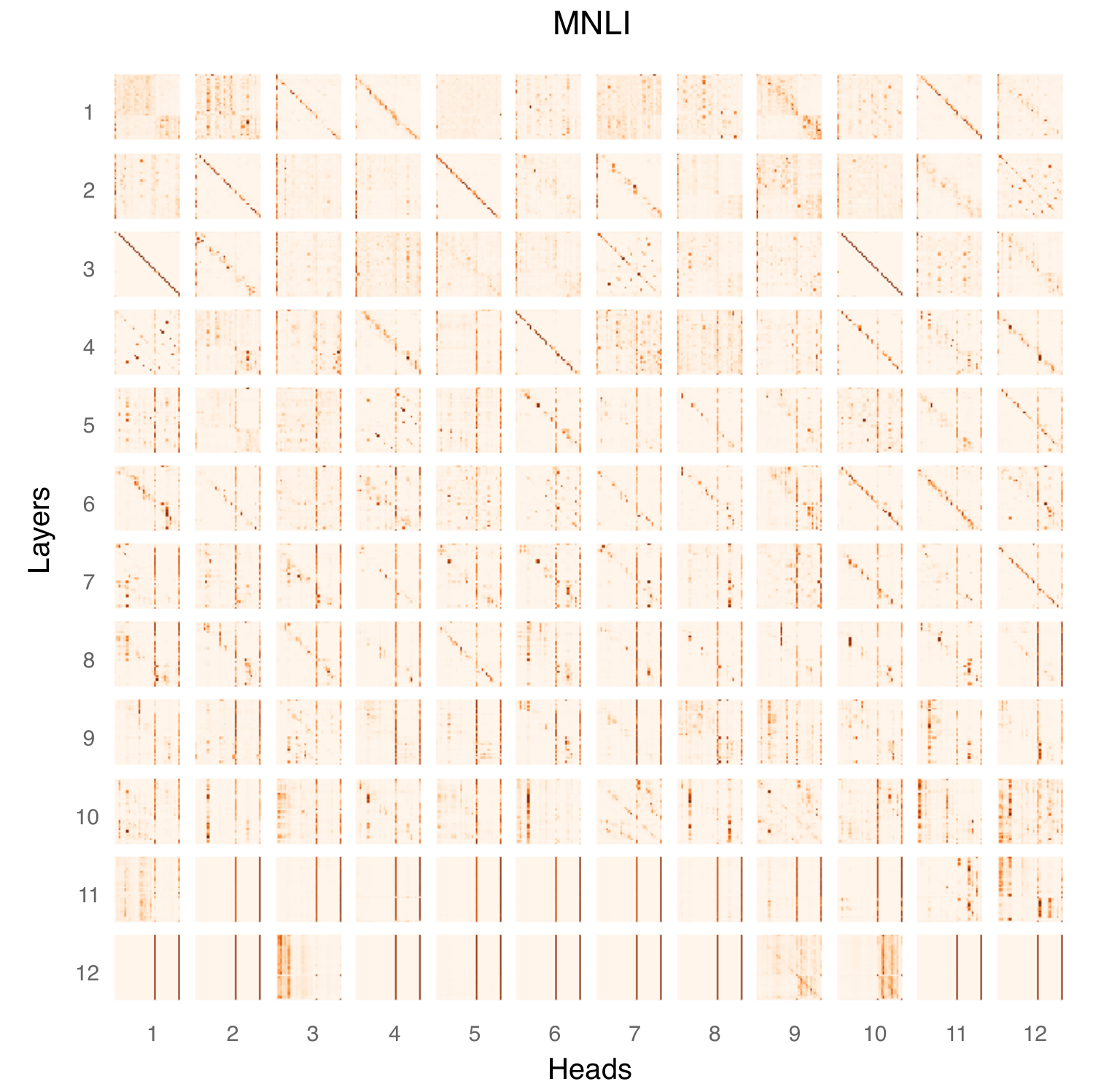}
\end{minipage}
\end{figure}

\label{sec:supplemental}

\end{document}